**Unified smoke and fire detection in an evolutionary framework with self-supervised progressive data augment**

Hang Zhang, Su Yang, Hongyong Wang, zhongyan lu, helin sun

**Abstract**

*Few researches have studied simultaneous detection of smoke and flame accompanying fires due to their different physical natures that lead to uncertain fluid patterns. In this study, we collect a large image data set to re-label them as a multi-label image classification problem so as to identify smoke and flame simultaneously. In order to solve the generalization ability of the detection model on account of the movable fluid objects with uncertain shapes like fire and smoke, and their not compactible natures as well as the complex backgrounds with high variations, we propose a data augment method by random image stitch to deploy resizing, deforming, position variation, and background altering so as to enlarge the view of the learner. Moreover, we propose a self-learning data augment method by using the class activation map to extract the highly trustable region as new data source of positive examples to further enhance the data augment. By the mutual reinforcement between the data augment and the detection model that are performed iteratively, both modules make progress in an evolutionary manner. Experiments show that the proposed method can effectively improve the generalization performance of the model for concurrent smoke and fire detection.*

## 1. Introduction

Fires are dangerous threaten to daily lives, and it is very urgent to develop effective fire detection systems. However, few endeavors have been devoted to the study of simultaneous detection of smoke and flame accompanying fires due to their different fluid natures and appearances, varying greatly with uncertain shapes, colors, textures, and motions. If taking into account the complex backgrounds of high variation, the problem becomes more challenging. Hence, in the previous studies, fire detection and smoke detection are treated as two independent tasks, that is, an independent detector should be designed and learnt for either task.

In order to realize simultaneous detection of smoke and flame, we treat fire detection as a multi-label image classification problem but most of the existing datasets do not label smoke and flame separately, so we collected a large image data set from the previous studies and redid the annotation with multiple class labels of smoke and flame attached to each image.

Due to the uncertain shape, size, and location of fluid objects, as well as the varying background in different scenarios, yet, the scenes with high variation from foreground to background impose severe challenge for fire detection in terms of generalization, even with the large-scale data at hand. The reason is: The limited training data cannot cover broad-spectrum scenarios as inclusive as possible in terms of enabling representative feature learning. The patterns learned from the training data by using a deep learner may not be extended to other scenes not presented in the training data. In such a context, a realistic solution is to apply data augment to enlarge the views of training by creating more scenes not existing in the original training set but helpful in granting generic representation learning.

In this study, we propose a random splicing data augment method by performing image stitch on a pair of positive and negative image examples to deploy resizing, reshaping, position variation, and background varying on the scenes such that the learner is enforced to be compatible with complex scenarios never seen before. Moreover, we propose a self-learning data augment scheme based on class activation map (CAM) [1], where the pixel-level activations with high confidence indicating the objects of interest form a highly trustable region and we incorporate it into the positive data set undergoing data augment. The data augment can thus be enhanced by incorporating such new data examples under the self-supervision by applying CAM, which leads to a new round of model training with better performance in general. The data augment with self-supervision and model training with newly augmented data are performed alternately to reinforce each other mutually until no improvement can be gained. Such an iterative learning scheme improves both components, namely, data augment and model training, incrementally.

The overall data augment can be divided into two stages: At the first stage, the model is trained only by using the random splicing data augment. Once the model has reached a certain degree of detection ability, we start the second stage self-learning to iterate self-supervision based data augment and model training alternately, where the data augment can strengthen and benefit from model training by applying and obtaining enlarged data set. After a couple of iterations, both data augment and model training can reach the best extent in terms of fire and smoke detection.

The contributions are summarized as follows:
(1) We collected and re-labeled a large-scale multi-label image dataset for classification of smoke and flame in fire detection.
(2) We propose a random splicing data augment method based on image stitch to deploy resizing, reshaping, position variation, and background changing to enable more generic and inclusive representation learning that is compatible with broad-spectrum scenarios.
(3) We propose a self-learning based data augment scheme to improve data augment and model training through mutual interaction in an iterative manner, by applying the new examples from CAM based self-supervision to the data augment module.
(4) Experiments show that the proposed method leads to better solution in terms of identifying smoke and fire simultaneously.

## 2. Related Works

### 2.1. Fire Detection

**Smoke detection:**

Smoke results from fire in general, but detection of smoke is much harder than detection of flame [2], since the rapid propagation caused motion and deformation make its fluid pattern appears with uncertainty. In contrast, the evolution of fire lasts for a longer period. So, most previous studies establish detector for either individual task.

For fire detection, early works were focused on feature engineering. Frequency features in wavelet domain were extracted based on the assumption that the scenes in surveillance videos are stationary [3]. A dual threshold AdaBoost algorithm with a staircase searching technique to detect smoke was proposed in [4].

The recent trend was shifted to apply deep learning to smoke detection. A spatiotemporal cross network (STCNet) was proposed to detect smoke in industrial scenarios [6]. A deep saliency network for frame-level and pixel-level smoke detection was proposed in [7].

Except for surveillance videos, photoelectric detectors and ionization detectors were also applied to detect smokes in fires [8]. However, the sensitivity, response time, reliability, and control area of deploying such non-visual sensors are not comparable to video surveillance due to the unpredictable propagation pattern of smoke.

**Flame detection:**

Many researches have been devoted to detect fires by means of temporal and spatial features of flames in video sequences. In [9], a video fire detection method is proposed by combining fast-RCNN with LSTM to extract spatial and temporal characteristics of flames. In [10], a video fire detection method is proposed by both considering the motion-flicker-based dynamic features and deep static features. In [11], a fire detection method used in urban nighttime environment is proposed: Firstly, the candidate areas are extracted using ELASTIC-YOLOv3, and then a histogram of the optical flow of the flame is extracted to get the dynamic characteristics. The work presented in [12] detects fire by utilizing optical flow, color, and motion features.

In [13], a pixel level fire detection method is proposed to convert the input image into a new color space that is easier to separate fire and non-fire pixels, and then apply the maximum inter-class threshold to get the pixels possibly corresponding with flame.

A lightweight fire detection model with high performance is proposed in [14], referred to as FireNet.

**Simultaneous smoke and flame detection:**

So far, only a few efforts have been made on simultaneous detection of smoke and flame in the context of computer vision. In [15], a color model is applied to detect smoke and flame simultaneously in fire. Besides, the object detection model, YOLO (you only look once), is used to detect smoke and flame [16].

In this study, we tackle fire detection as a multi-label image classification problem with simultaneous detection of both smoke and flame.

### 2.2. Data augment methods

For image data, the widely used data augment methods include graying, flipping, clipping, color adjustment, size scaling, and brightness adjustment.

We sort the recent data augment methods into two categories: Editing the original image only and combining multiple images data.

**Editing the original image only:**

Cutout [17] and random erasing [18] randomly cut off some image blocks in the original image, including regular rectangular blocks and irregular blocks, so as to generate occlusion areas, and then fill them with blank or noise. However, these methods may block or leave the object out when it is small. AutoAugment [19] proposed to apply automatic search such as reinforcement learning to train the model, so as to learn the optimal strategy for data augment. KeepAugment [20] proposed to incorporate a salient region judgment method: Firstly, the salient region is detected to generate a salient map such that the most salient region in the image can be retained during data augment, avoiding the problem of leave the salient region out of the image when applying cutout and random erasing.

However, saliency detection is also challenging. Recently, an open-source Python code referred to as "imgaug" [21] was proposed to impose effects like noise, blur, pooling, distortion, geometric transformation, and weather effect.

**Combination of multiple images:**

MixUp [22] performs pixel-level linearly weighted sum across two images from different categories, and the data labels are then smoothed by the weights. Mixed-Example [23] proposes an improved scheme for MixUp, that is, the multiple images are weighted with nonlinear functions, so as to obtain a variety of augmented data. Similar to Cutout, CutMix [24] also cuts off some image blocks from images in the training set but uses other types of image data to fill the cut-off areas. At the same time, the class label is calculated according to the ratio of the areas of the two combined images. RICAP [25] proposed a data augment method of random clipping and stitching: Firstly, multiple images are randomly selected, and the clipping size of each image is calculated to make the combined image following clipping and stitching possess the same size as that of the original image. Then, clipping and stitching are performed. This method also uses the ratio of the areas preserved for either image to smooth the class label. In the problem of object detection, [26][27] proposed a method by segmenting the target object, and then pasting the segmented object on random background.

3. **Multi-label image dataset for fire-related smoke and flame classification**

In view of the problem that smoke and flame are rarely annotated separately in previous fire detection datasets, we collect a couple of such datasets and redo the annotation on fire and smoke separately, including Fire-Detection-Image-Dataset [28], datasets used in [29], RF dataset and SF dataset [30], Fire Image Data Set for Dunnings 2018 study - PNG still image set [31], Kaggle Fire Detection Dataset [32], and fire dataset [9]. In addition, in order to provide sufficient negative examples, we also collected some non-smoke and non-fire datasets, including Road CCTV images with associated weather data [33], LIVE Image Defogging Database [34], BeDDE [35], exBeDDE [36], Landscape Dataset [37], and Barcelona Dataset [38]. At the same time, we also collected some fire video datasets, including FIRESENSE database of videos for flame and smoke detection [39], Fire Detection Dataset [40], fire dataset [9], smoke videos dataset used in [41], videos dataset used in [7], where the video frames are sampled to avoid too much meaningless duplicate data.

After eliminating some duplicate data and the data whose category is difficult even for human eyes to judge, we obtained a large-scale multi-label image classification dataset of fire-related smoke and flame by manually annotating one by one. Statistics of each part of the dataset are shown in Table 1. It is worth mentioning that there are many negative samples that are very similar to smoke or flame, such as light, sunset glow, clouds, haze, and water vapor. We refer to these data as "difficult negative samples" and have them form an independent class, say, negative set of complex scene in Table 1, while the others are referred to as "simple negative samples", namely, negative set of simple scene in Table 1. The representative examples are shown in Figure 1, where each line from top to bottom corresponds with simple negative, difficult negative, smoke-only, fire-only, and smoke and fire examples, respectively.

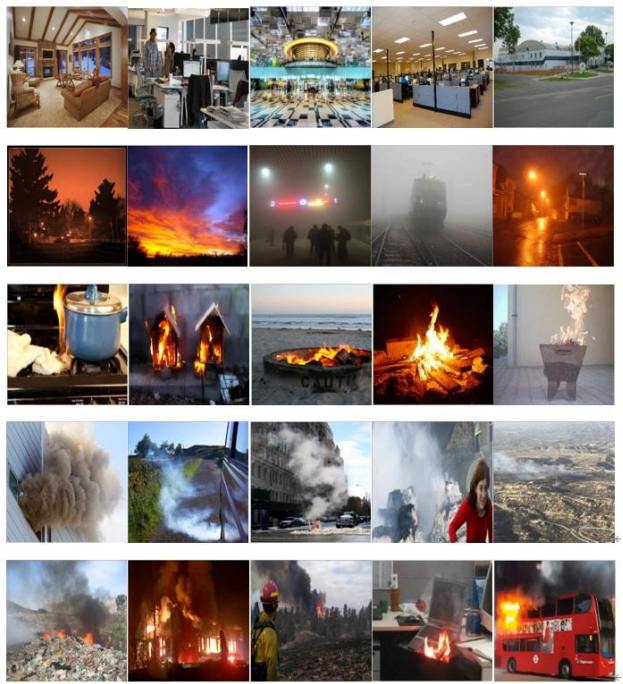

Figure 1: The representative examples in the multi-label image dataset of fire-related smoke and flame, where each line from top to bottom corresponds with simple negative, difficult negative, smoke-only, fire-only, and smoke and fire samples, respectively.

| Positive set (#17622) | | | Negative set (#45131) | |
|---|---|---|---|---|
| Fire only | Smoke only | Fire and smoke | Simple scene | Complex scene |
| #3294 | #4767 | #9561 | #34188 | #8817 |

Table 1: Statistics of the multi-label image dataset of fire-related smoke and flame.

During training, we only use the data shown in Table 1, where 80% data from each category forms the training set, and the rest serves as the verification set.

For the testing data, State Grid Beijing Electric Power Company provided 13 surveillance videos for fire monitoring in real-world scenarios. Most of them are very ambiguous, where the frames of 5 videos are close to gray images, imposing a severe challenge for the detection problem. We label each individual video frame to form test dataset 1, the statistics of which is shown in Table 2.

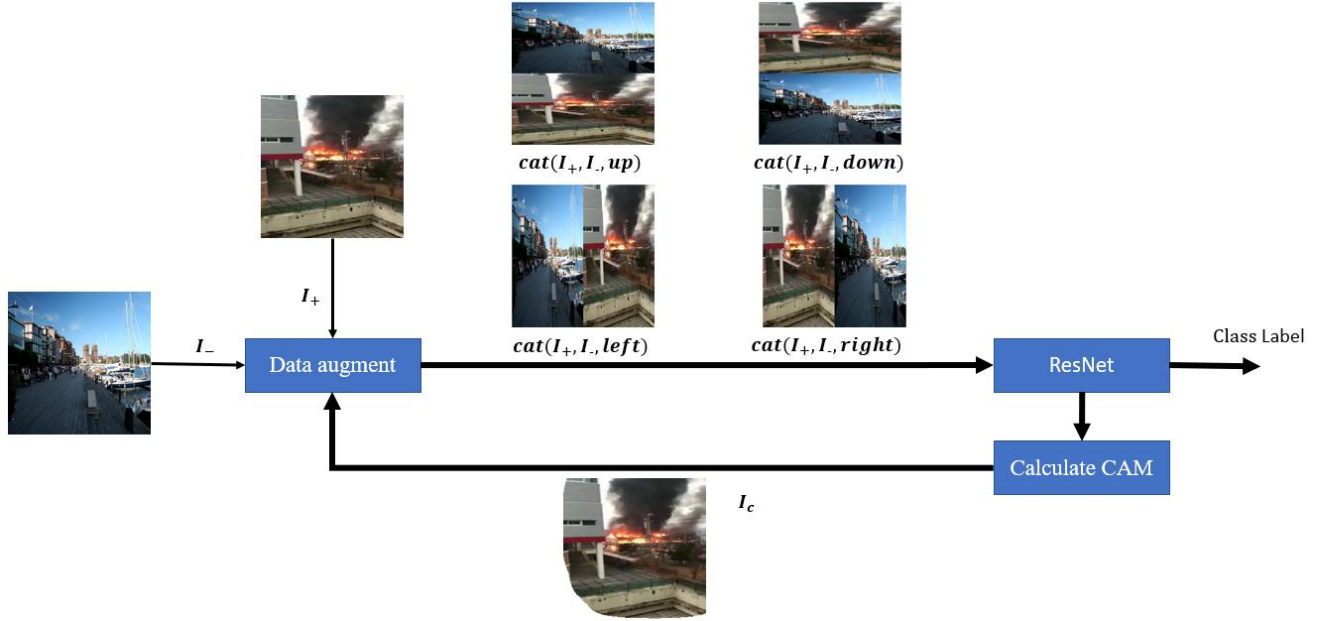

Figure 2: Pipeline of the proposed self-learning data augment framework.

| Positive set (#11206) | | | Negative set (#33600) |
|---|---|---|---|
| Fire only | Smoke only | Fire and smoke | |
| #409 | #5980 | #4817 | #33600 |

Table 2: The test dataset 1.

Then, we collected in total 7 news videos and surveillance videos from Internet as fire and non-fire to form test dataset 2, and labeled these data similarly, the statistics of which is shown in Table 3.

| Positive set (#11206) | | | Negative set (#33600) |
|---|---|---|---|
| Fire only | Smoke only | Fire and smoke | |
| #7 | #4325 | #4260 | #5127 |

Table 3: The test dataset2.

## 4. Methodology

We use ResNet34 [42] as the backbone of our detector for image classification, where only the last fully connected layer is modified as two neurons to predict the confidence of smoke and fire.

As shown in Figure 2, the proposed self-learning data augment framework can be divided into two stages: At the first stage, the model is trained only by random splicing data augment method, based on which the model has reached its best extent to grant the detection accuracy after a period of training. To further improve its discriminating power, we then begin the second stage of training to gradually augment the data in the way of self-learning. At this stage, the model learning and data augment are mutual reinforced in an iterative manner, forming a closed loop. In detail, the detection model is accompanied by a parallel module, namely, class activate map (CAM) [1], which is the byproduct of the classifier, highlighting the importance of each image pixel in terms of contributing to the classification by forming the pixel-level confidence as a heat map so as to figure out possible existence of targets. Under the supervision of CAM, we can segment the highly reliable regions enclosing possibly the targets of interest of positive samples, and apply them to the next round data augment to further strengthen the training of the classification model.

In the following, we first introduce the random splicing data augment method, and then the self-supervised data augment as well as the mutually interacted evolution framework for model enhancement in detail.

### 4.1. Random splicing data augment method

Based on the fact that shape plays a minor role in discriminating smoke and fire as they are fluid objects, we proposed a random splicing data augment method without preserving shapes.

As shown in algorithm 1, the random splicing data augment method includes the following key steps: (1) A threshold $\theta$ is set at first to determine the probability of performing data augment. (2) For the input positive example undergoing data augment, n≥0 simple negative samples are randomly selected, each of which forms a stitched image with the positive one. The reason to perform data augment with the simple negative samples only is to prevent the model from learning features from the difficult negative samples that are too ambiguous to be distinguished from true positive ones. (3) As shown in Figure 2, the direction of the image stitch to generate new data examples can randomly be any direction of up, down, left, and right. (4) The image pair to perform data augment should be adjusted to the same size for collage.

**Algorithm 1: Data augment**

**Input:**
$I_+$: Positive example
{$I_-$}: Negative examples
$I_c$: Output of algorithm 2
$\theta$: Threshold to perform data augment
N: Maximum number of negative samples applied
**Output:**
{$I_{aug}$}: List of stitched image

**Algorithm:**
If random(0,1)<$\theta$, return;
If $I_c$ exists
- Perform random scaling on $I_c$ to obtain $S(I_c)$
- Paste $S(I_c)$ onto $I_-$ at random position to obtain $I_{c+}$=overlap($S(I_c), I_-$);
- Replace $I_+$ with $I_{c+}$;

End if
n= random(0, N);
While n>0:
- Read a negative example $I_-$ at random;
- $I_+$=resize($I_+$);
- $I_-$=resize($I_-$);
- $dir$=random(up, down, left, right);
- Add $I_{aug}$=cat($I_+, I_-, dir$) to the output list;
- n-1→n;

End while

The motivation of the data augment is as follows: (1) Since shape varies greatly for a fluid object such as fire or smoke, the image stitch-based data augment can result in deformation, scaling, and translation of the fluid objects so as to enlarge the data set in incorporating more complex patterns to improve the discriminating power of the detector. Especially, the resizing of smoke and fire will be helpful in learning the features of small objects. (2) Background is also a key factor to affect detection performance. The randomly selected negative examples undergoing data augment can create more diverse background scenarios in the present of the same foreground, enlarging the view of the model to be aware of the same semantics over more possible backgrounds.

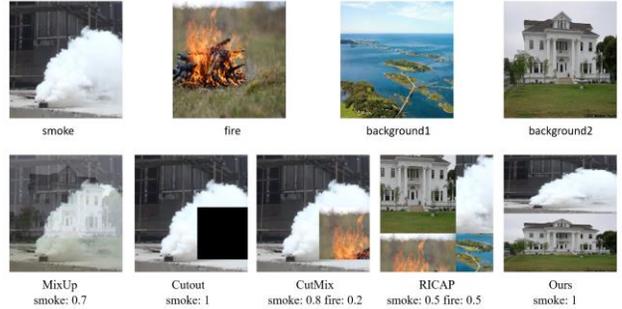

Figure 3: An example of comparison between some well-known methods and the proposed method in data augment.

As shown in Figure 3, we simply implement some of the well-known data augment methods to enable an intuitive comparison with the proposed random splicing data augment method. The first row includes a smoke sample, a flame sample, and two negative samples. The second row are the results from MixUp, cutout, CutMix, RICAP, and the proposed method. It can be seen that our method is similar to CutMix and RICAP, but we believe that CutMix and RICAP may be more suitable for such a scenario that the object is big enough to occupy almost the whole image, since it is very likely to lose the foreground object during clipping. Our method is more suitable for fluid objects with uncertain shapes, sizes, and positions since the proposed data augment scheme create various fluid patterns by peforming resizing, reshaping, position variation, and background changing.

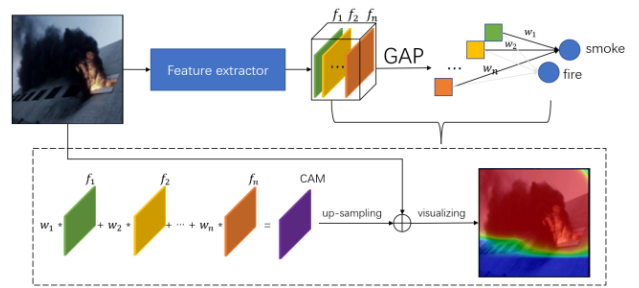

Figure 4: Schematic diagram of the CAM[1] calculation process.

### 4.2. Self-supervised progressive data augment

The schematic diagram of the CAM calculation process is shown in Figure 4, where $f_1$ to $f_n$ represent the feature maps obtained by the last convolution layer, and $w_1$ to $w_n$ represent the weights in the fully connected network for classifying the smoke feature vectors obtained from the global average pooling (GAP) output of the feature maps $f_1$ to $f_n$, and the weighted sum of these values is smoke CAM. We use bilinear interpolation method to up sample the CAM to obtain the smoke confidence of each position of the image as the final smoke CAM. Then we visualize it by linear weighting with the original image, and it can be seen that CAM of smoke can well highlight the area of smoke. The fire CAM can be calculated in the same way.

The key steps of the self-supervised progressive data augment method include: (1) Calculate the CAM for the positive sample to obtain pixel-level activation degree as a heat map, and the highly reliable area possibly containing the objects of interest is segmented by looking up at the pixel associated confidence value greater than 0 both in smoke CAM and fire CAM. (2) The highly trusted region output from CAM is pasted onto the randomly selected negative sample at a random position following random scaling in terms of height and width. (3) The image obtained by step 2 is applied to algorithm1 as positive example to undergo further data augment.

As shown in algorithm 2, the model training is redone based on the self-supervised progressive data augment method as follows: (1) Initialize the model with the parameters learnt from the previous stage of training. (2) Use the data output from algorithm 2 to augment the training data, and train 100 epochs for the model. (3) During verification, the self-supervised progressive data augment method is also applied to augment the data in the verification set, and the model with the lowest loss in verification is preserved with its parameters. (4) If the loss of this iteration is lower than that of the previous iteration, continue the next iteration. Otherwise, end the training process.

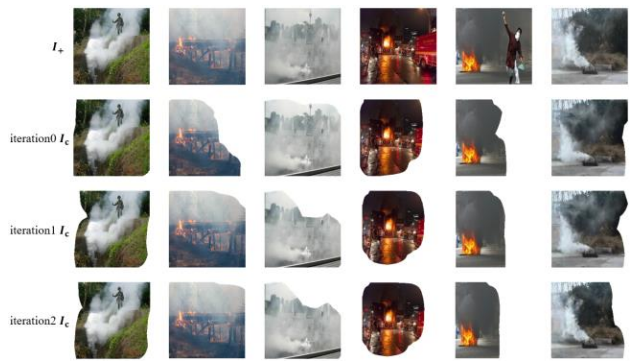

Figure 5: Visualization of the high confidence region of some images generated in each iteration of Algorithm 2.

be seen that the high reliability region segmented by CAM is evolving.

## 5. Experiments

The implementation is based on PyTorch, and the experiments are carried out on a server equipped with NVIDIA GTX1080Ti graphics card. The model used for fire/smoke detection is the modified ResNet34 [42], and the loss function is the Binary Cross Entropy Loss (BCE Loss), widely used in multi-label image classification problems. In the training process, we applied the data augment method to both the training set and the verification set, but we did not augment the test data so as to facilitate the comparison of the experimental results.

**Algorithm 2: Self learning**

**Input:**
- $\{I_{aug}\} \cup \{I_+\} \cup \{I_-\}$
- **Model**: Model with the parameters learnt from the first stage of training

**Output:**
- **Model**: Model trained on $\{I_{aug}\} \cup \{I_+\} \cup \{I_-\}$
- $I_c$: Highly trustable area resulting from CAM

**Algorithm:**
- Train **Model** with 100 epochs, and preserve the solution with the minimum loss on the validation set;
- If the loss is no longer decreases in two consecutive rounds of training, quite;
- Compute $I_{cam}$=CAM($I_+$), where CAM is the byproduct of **Model**;
- Retrieve the area of $I_{cam}$>0 to obtain $I_c$;
- Call algorithm 1 to update $I_{aug}$ with $I_c$;
- Call algorithm 2 to update **Model** and $I_c$;

| Operation | Probability |
|---|---|
| **The proposed method** ($\theta$ in Algorithm 1) | 0.5 |
| **Grayscale convert** | 0.01 |
| **Flip vertically** | 0.02 |
| **Flip horizontally** | 0.5 |
| **Rotation ±30°** | 0.3 |
| **1±0.3 times brightness** | 0.5 |
| **1±0.3 times contrast** | 0.5 |
| **0.2~1.6 times saturation** | 1 |
| **1±0.01 times hue** | 0.1 |

Table 4: Data augment methods and parameter setting

Before applying the data into the model, we resize all the images to the size of 224 pixels wide and 224 pixels high. In algorithm 1, the height and width of $I_c$ is randomly scaled to an interval of [100, 224]. Besides, all the data underwent Z-score standardization prior to being applied, where the mean and variance of each channel are set to 0.5 as hyper parameters. For the data augment without CAM, we set the learning rate to 0.0004. Otherwise, the learning

As shown in Figure 5, we visualize the high confidence region of some images generated in each round, and it can rate is set to 0.0000001. Batch size is set to 200. In addition to the random splicing data augment, we also use some widely used traditional data augment methods, such as rotation, flip, and change of brightness, contrast, saturation, and hue. Whether to apply these methods to each data example is determined according to a certain probability as defined in Table 4.

### 5.1. Performance

Some well-known classification methods including VGG19 [43] and ResNet34[42], the widely used target detection algorithm YOLOv5 [44], a newly proposed fire detection method referred to as FireNet [14], and the traditional method Optimal Color Space (OCS) [13] are compared with the proposed method, where the best result obtained by each model are compared.

The datasets used in the comparative experiment are test dataset 1 and test dataset 2. Here, we use the true positive rate (TPR), false positive rate (FPR), and Area under Curve (AUC) to evaluate the method, among which we set the confidence threshold to 0.5 to compare the TPR and FPR. The results on test set 1 and test set 2 are shown in Table 5 and Table 6, respectively.

| Method | Smoke TPR | Fire TPR | Smoke FPR | Fire FPR | Smoke AUC | Fire AUC |
|---|---|---|---|---|---|---|
| VGG19 | 0.109 | 0.195 | 0.054 | 0.030 | 0.841 | 0.859 |
| ResNet34 | **0.398** | 0.578 | 0.117 | 0.008 | 0.735 | 0.961 |
| YOLOv5 | 0.372 | 0.048 | 0.061 | **0.000** | 0.771 | 0.797 |
| OCS | | 0.160 | | 0.345 | | |
| FireNet | 0.170 | 0.104 | 0.068 | **0.000** | 0.681 | 0.966 |
| Ours | 0.133 | **0.800** | **0.003** | 0.003 | **0.848** | **0.988** |

Table 5: Comparison on test dataset 1

| Method | Smoke TPR | Fire TPR | Smoke FPR | Fire FPR | Smoke AUC | Fire AUC |
|---|---|---|---|---|---|---|
| VGG19 | 0.586 | 0.748 | 0.003 | 0.018 | 0.974 | 0.942 |
| ResNet34 | 0.697 | 0.767 | 0.007 | **0.009** | 0.974 | 0.959 |
| YOLOv5 | 0.113 | 0.535 | 0.003 | 0.028 | 0.900 | 0.865 |
| OCS | | 0.626 | | 0.576 | | |
| FireNet | 0.615 | 0.645 | 0.015 | 0.014 | 0.974 | 0.940 |
| Ours | **0.758** | **0.910** | **0.003** | 0.028 | **0.983** | **0.986** |

Table 6: Comparison on test dataset 2

It can be seen that although ResNet34 outperforms the proposed method in terms of smoke TPR on test dataset 1, but at the cost of leading to increased smoke FPR as high as 0.117. Besides, although the fire FPR obtained by FireNet is much lower than that of the method proposed on test dataset 1, its performance on TPR is relatively low. In addition, the proposed method performs better on test set 2, except for the fire FPR. According to AUC, the proposed method performs the best in an overall sense on both test sets.

### 5.2. Ablation study

We use ResNet34 [42] as the backbone of our detector to evaluate the contribution of the proposed data augment method. As shown in Table 7 and table 8, ResNet34 only uses the traditional data augment method to train the model without the random splicing data augment and the self-supervised progressive data augment. Moreover, after the first stage of training, we only train the model by adjusting the learning rate to 0.0000001 without adding the self-learning data augment method to test its impact.

We use AUC only to evaluate how every component affects a solution in various configurations. The results are shown in table7 and table8, respectively.

It can be seen that including the self-supervised progressive data augment at stage2 leads to the optimal solution in terms of AUC, demonstrating the add value the self-supervised progressive data augment in comparison with using the random splicing only. Besides, ResNet34 without the random splicing and self-learning data augment performs the worst, which means that removal of the random splicing data degrades the solution.

| Method | Smoke AUC | Fire AUC |
|---|---|---|
| ResNet34 | 0.73484 | 0.96080 |
| Stage1 | 0.79391 | 0.96405 |
| Stage1+ Adjust learning rate | 0.80323 | 0.98165 |
| Stage2 | **0.84808** | **0.98764** |

Table 7: Ablation study on test dataset1

| Method | Smoke AUC | Fire AUC |
|---|---|---|
| ResNet34 | 0.97370 | 0.95890 |
| Stage1 | 0.97748 | 0.96704 |
| Stage1+ Adjust learning rate | 0.97226 | 0.98454 |
| Stage2 | **0.98288** | **0.98612** |

Table 8: Ablation study on test dataset2

### 6. Conclusion

Aiming at the challenging problem of fire and smoke detection, we advance the literature by realizing simultaneous detection of smoke and fire. We contribute multi-label annotations on a large data set collected from a couple of previous studies and demonstrate the effectiveness and generalization ability of the model using heterogeneous data sources. In terms of data augment, we propose an image stitching based method to deploy resizing, deforming, position variation, and background altering. Based on class activation map, the highly trustable region under self-supervision is applied as new source of positive examples to enhance the data augment, by which the model training and data augment are mutually reinforced in an evolutionary manner.